%% file: main.tex
\documentclass[10pt,twocolumn,letterpaper]{article}

\usepackage{cvpr}              


\definecolor{cvprblue}{rgb}{0.21,0.49,0.74}
\usepackage[pagebackref,breaklinks,colorlinks,allcolors=cvprblue]{hyperref}
\usepackage{array,multirow}

\def\paperID{6} 
\def\confName{CVPR}
\def\confYear{2025}

\title{Efficient 2D to Full 3D Human Pose Uplifting including Joint Rotations}

\author{Katja Ludwig$^*$, Yuliia Oksymets$^*$, Robin Schön, Daniel Kienzle, \& Rainer Lienhart\\
Chair for Machine Learning \& Computer Vision, University of Augsburg, Germany\\
{\tt\small\{firstname.lastname\}@uni-a.de} \\
}

\begin{document}
\maketitle
\def\thefootnote{*}\footnotetext{Equal contribution}\def\thefootnote{\arabic{footnote}}
\input{sec/0_abstract}    
\input{sec/1_intro}
\input{sec/2_related_work.tex}
\input{sec/3_method}
\input{sec/4_experiments}
\input{sec/5_conclusion}

{
    \small
    \bibliographystyle{ieeenat_fullname}
    \bibliography{main}
}


\end{document}

%% file: sec/0_abstract.tex
\begin{abstract}
In sports analytics, accurately capturing both the 3D locations and rotations of body joints is essential for understanding an athlete's biomechanics. While Human Mesh Recovery (HMR) models can estimate joint rotations, they often exhibit lower accuracy in joint localization compared to 3D Human Pose Estimation (HPE) models. Recent work \cite{a2b} addressed this limitation by combining a 3D HPE model with inverse kinematics (IK) to estimate both joint locations and rotations. However, IK is computationally expensive.
To overcome this, \textbf{we propose a novel 2D-to-3D uplifting model that directly estimates 3D human poses, including joint rotations, in a single forward pass}. We investigate multiple rotation representations, loss functions, and training strategies — both with and without access to ground truth rotations. Our models achieve state-of-the-art accuracy in rotation estimation, are 150 times faster than the IK-based approach, and surpass HMR models in joint localization precision.
\end{abstract}

%% file: sec/1_intro.tex
\vspace{-0.1cm}
\section{Introduction}
\label{sec:intro}

Classical monocular 3D Human Pose Estimation (HPE) methods have shown impressive results in recent years. They estimate a 3D human pose consisting of a set of 3D keypoints and a skeleton from either a single image or a video. Most promising methods are 2D to 3D uplifting methods, meaning that they first estimate 2D keypoints in each frame of a video and then lift them to 3D via an upsampling model operating on 2D pose sequences.      
In sports, a significant limitation of estimated 3D poses from such models is that they do not capture the rotation of body parts. However, rotations are crucial for sports analytics, as they are essential for understanding an athlete's biomechanics and calculating the forces and torques acting on the body.

\begin{figure}[htb]
    \centering
    \includegraphics[width=\linewidth]{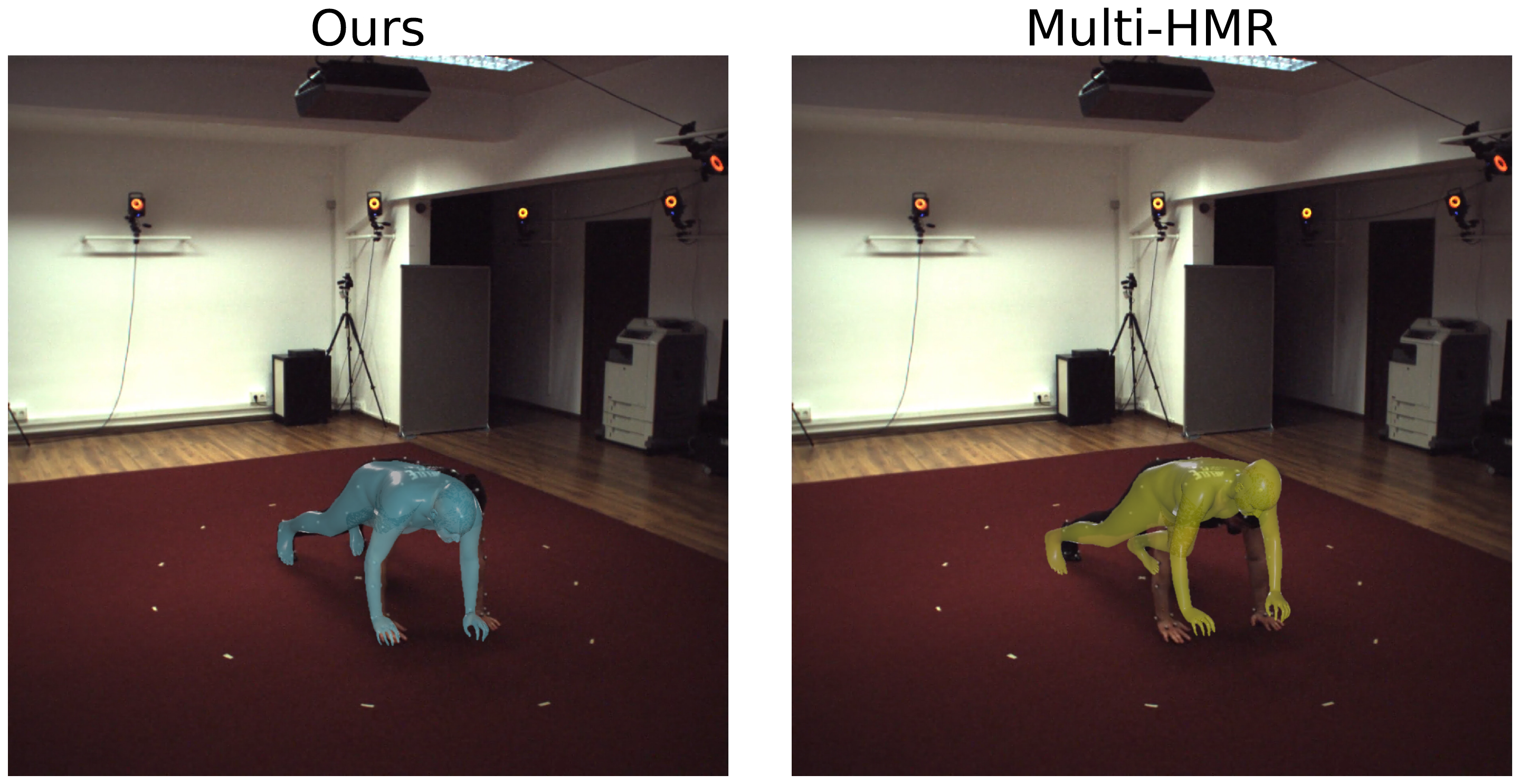}
    \includegraphics[width=\linewidth]{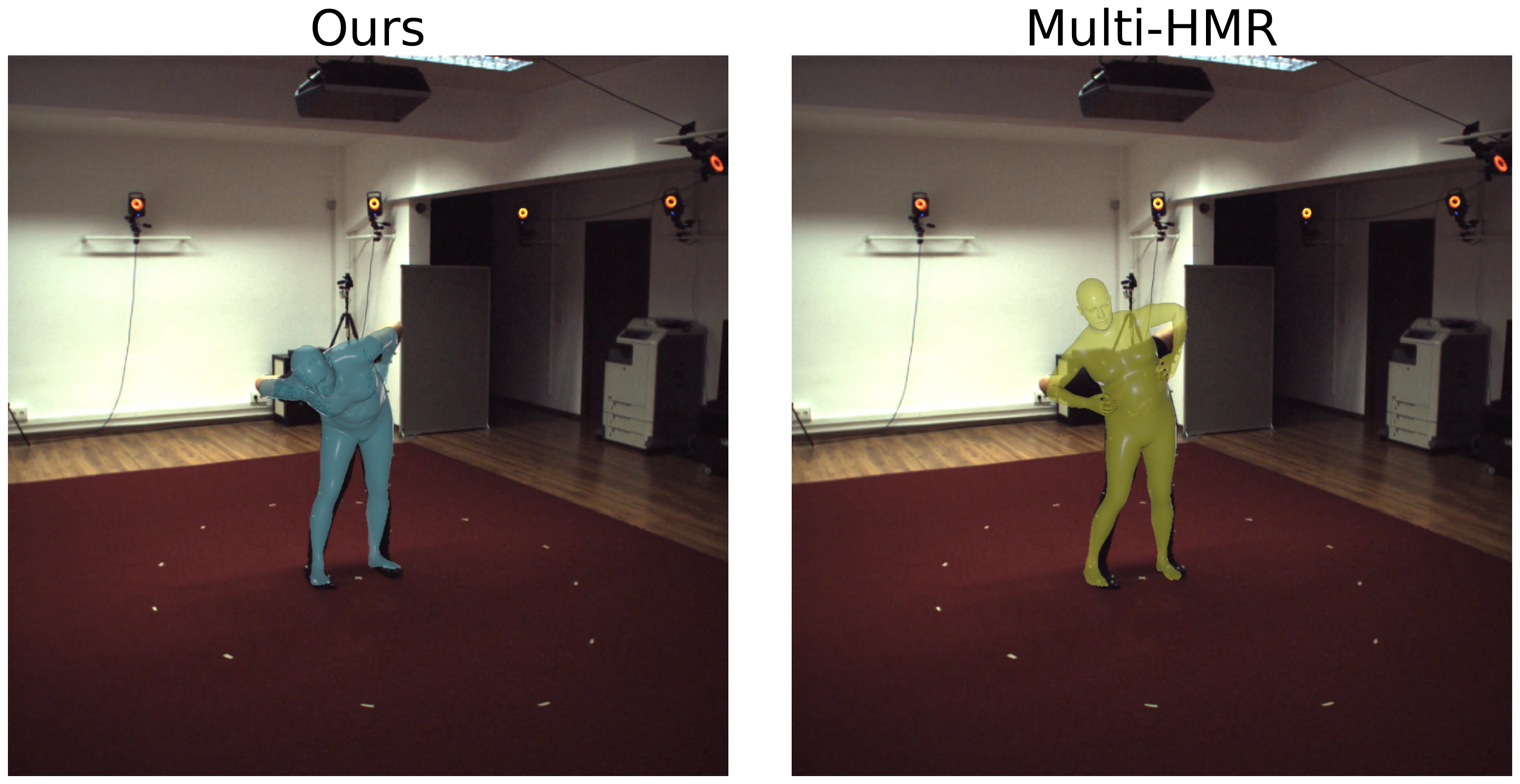}
    \vspace{-0.6cm}
    \caption{Two examples comparing the results of our model (blue, left column) compared to those of the SOTA HMR model Multi-HMR \cite{multihmr} (yellow, right column). We display meshes created with the estimated rotations and the ground truth body shape.}
    \label{fig:vis}
    \vspace{-0.4cm}
\end{figure}

In contrast, Human Mesh Recovery (HMR) models estimate 3D human meshes. Most of these models rely on a parametric representation of the human body, such as SMPL-X \cite{smplx}, which explicitly separates body shape and pose. The body shape is modeled as a low-dimensional embedding, while the body pose is defined by a set of 3D joint rotations. A mesh in a template pose is generated by applying the shape parameters to the parametric model, from which joints are regressed. The estimated joint rotations are subsequently applied to the regressed joints and the corresponding body parts to create the final posed mesh. 
As a result, HMR models based on parametric body models like SMPL-X can recover full 3D poses, including joint rotations. However, they have a key limitation: They are not able to leverage temporal information from long video sequences.
Especially in the field of sports with high-speed movements and extreme poses, this leads to worse detection accuracy regarding keypoint locations compared to 3D HPE models, as investigated by Ludwig et al. \cite{a2b}. Therefore, they propose to combine a 3D HPE model with a body shape estimation model and inverse kinematics (IK) to recover a human mesh with more accurate joint locations. They need to apply IK to the 3D poses estimated by the 3D HPE model to obtain the joint rotations, since they are not included in the output of 3D HPE models. IK is a complex and computationally expensive process, since it is an optimization-based approach required to run for each frame.

In this paper, we propose a different approach, which estimates the full 3D pose, including rotations. We extend a recent 2D to 3D uplifting model such that it can estimate a 3D pose including the rotations. Apart from the precision of the estimated joints, we further evaluate the precision of the estimated rotations and show that our model outperforms other SOTA models regarding the accuracy of the estimated rotations. For sports analysts, these rotations are crucial for accurately analyzing an athlete's biomechanics. Furthermore, our proposed model outperforms Ludwig et al. \cite{a2b} in speed, as it eliminates the need for an additional IK step.
Our contributions can be summarized as follows: \footnote{The code is available at \url{https://github.com/kaulquappe23/full_3d_hpe_uplifting}}

\begin{itemize}
    \item We introduce a novel 2D to 3D uplifting model capable of \textbf{estimating 3D human poses, including joint rotations}. We explore multiple rotation representation variants and compare models trained with and without direct supervision on joint rotations.
    
    \item A comprehensive evaluation of both joint position and rotation accuracy demonstrates that \textbf{our model achieves superior rotation estimation performance} compared to existing SOTA approaches. 
    
    \item Our proposed models offer a substantial \textbf{improvement in computational efficiency} over the method of Ludwig et al. \cite{a2b} by eliminating the need for an additional IK step, while maintaining comparable joint position accuracy and enhancing rotation estimation.
\end{itemize}

\begin{figure*}[t]
    \centering
    \includegraphics[width=0.9\linewidth]{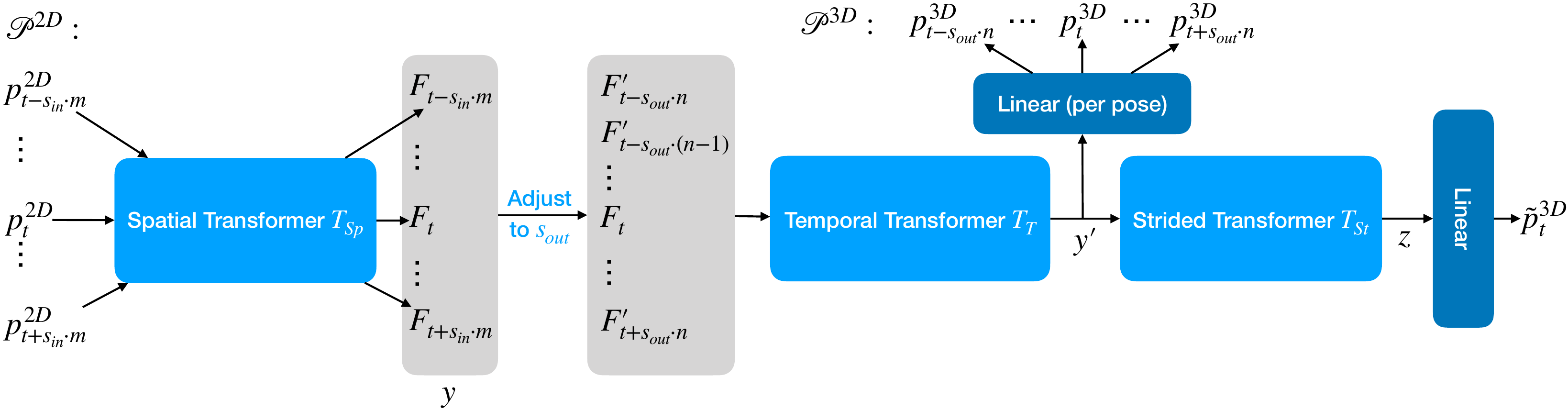}
    \caption{General architecture of the UU model. A pose sequence is fed through an intra-pose operating spatial Transformer $T_{Sp}$ followed by an inter-pose operating temporal Transformer $T_{T}$. A linear layer outputs 3D pose estimates for each pose in its input sequence, while a strided Transformer $T_{St}$ with a final linear layer reduces the sequence length to output a single 3D pose estimate for the central frame at position $t$ in the input sequence.}
    \label{fig:uu}
    \vspace{-0.4cm}
\end{figure*} 

%% file: sec/2_related_work.tex
\section{Related Work}

\textbf{2D to 3D Pose Uplifting.} To improve 3D joint localization, recent works have leveraged context from neighboring frames in videos. Pavllo et al. \cite{Pavllo_2019_CVPR} propose an uplifting model based on a temporal convolutional network (TCN), which processes long input sequences and models local context by convolving neighboring frames. 
To model spatial and temporal correlations simultaneously across joints, subsequent works \cite{Cai_2019_ICCV, hu2021conditional} utilize graph convolutional networks. Recently, Transformer-based architectures have become popular for capturing spatio-temporal correlations. PoseFormer \cite{Zheng_2021_ICCV} stacks a temporal Transformer to learn global dependencies between frames and a spatial Transformer to capture local joint correlations. Li et al. \cite{li2022exploiting} leverage a strided Transformer to efficiently process long input sequences. We select the Uplift and Upsample (UU) model \cite{uu} as the backbone architecture for our models because it is a very efficient SOTA 3D HPE model. It combines spatial, temporal, and strided Transformers. Ludwig et al. \cite{a2b} combined UU with IK to estimate joint rotations, but their approach is computationally expensive.

\textbf{Human Mesh Recovery.} HMR has been an active area of research in the last years. The parametric SMPL-X \cite{smplx} body model disentangles the parameter set for pose and shape and has established a stable foundation for HMR. Cai et al. \cite{cai2023smpler} design a Vision Transformer based generalist HMR foundation model using 4.5M training examples from diverse data sources. The challenges caused by smaller features, such as hands and facial expressions, have lead to numerous works that use multi-crop pipelines \cite{feng2021collaborative, Moon_2022_CVPR, choutas2020monocular}. 
Recent approaches extend HMR to multi-person settings, where two-stage pipelines with a human detector and a single-person mesh estimation model dominate \cite{Choi_2022_CVPR, goel2023humans}. %
Sun et al. \cite{sun2022putting} propose a single-shot network with an imaginary Bird’s-Eye-View to efficiently reason about depth in a multi-person setting. 
Qiu et al. \cite{qiu2023psvt} leverage Transformers to capture spatio-temporal context among instances in an end-to-end manner. 
For comparison with our proposed approaches, we select the SOTA model Multi-HMR \cite{multihmr}, which is a single-shot, multi-person Transformer-based network building upon the works of Sun et al. and Qiu et al. \cite{sun2022putting, qiu2023psvt}.

\textbf{Learning with Rotations.} Many computer vision tasks, such as pose estimation from images \cite{do2018deep, xiang2017posecnn} and point clouds \cite{gao2018occlusion} as well as structure from motion, perform regression on rotations \cite{Ummenhofer_2017_CVPR}. Evaluating the distance between two 3D rotations is often an essential task. Many works use axis-angle vectors or quaternions to represent 3D rotations. In early work, Huynh et al. \cite{huynh2009metrics} argue that quaternions are the most efficient representations both spatially and computationally and propose several distance metrics for different representations. Levinson et al. \cite{levinson2020analysis} demonstrate that symmetric orthogonalization of rotation matrices via SVD achieves SOTA performance. Zhou et al. \cite{zhou2019continuity} discourage using 3D or 4D representations, as they introduce discontinuities in the optimization process. 

%% file: sec/3_method.tex
\section{Method}

\textbf{Base Model.} All our model variants are based on the Uplift and Upsample (UU) architecture proposed by Einfalt et al. \cite{uu}. We briefly recap its architecture, which is visualized in Figure \ref{fig:uu}. As an input, the UU model takes a sequence of 2D poses $\mathcal{P}^\text{2D} = p^\text{2D}_{t-s_\mathit{in}\cdot m}, ..., p^\text{2D}_{t+s_\mathit{in}\cdot m}$ around a central frame $p^\text{2D}_{t}$ at time $t$. Special for the UU model is that this sequence has a stride, hence the poses are not of subsequent frames, but are spaced apart by a fixed number of frames, which is the reason for its efficiency. At first, a spatial Transformer $T_{Sp}$ is applied to each 2D pose separately to enhance the pose-internal representation. This results in a sequence $y$ of enhanced feature tokens per pose. If the output stride $s_\mathit{out}$ is lower than the input stride $s_\mathit{in}$, which is used to efficiently generate a denser output, these tokens are padded with special upsampling tokens for every missing frame in the sequence. Then, they are fed through a temporal Transformer $T_T$, which operates across the pose tokens. A linear layer is applied to every pose token from the output sequence $y'$, resulting in an auxiliary output sequence of 3D poses $\mathcal{P}^\text{3D} = p^\text{3D}_{t-s_\mathit{out}\cdot n}, p^\text{3D}_{t-s_\mathit{out}\cdot (n - 1)}, ..., p^\text{3D}_{t+s_\mathit{out}\cdot n}$ with the output stride $s_\mathit{out}$. Last, $y'$ is fed through a strided Transformer $T_{St}$, which gradually reduces the sequence length and outputs a single enhanced 3D pose $\tilde{p}^\text{3D}_{t}$ for the central frame of the input sequence. The final output is $\tilde{p}^\text{3D}_{t}$. A root-relative mean per-joint position error (MPJPE) is used as the loss function $L_\mathit{joint}$ for both $\tilde{p}^\text{3D}_{t}$ and the auxiliary output sequence $p^\text{3D}_{t-s_\mathit{out}\cdot n}, ..., p^\text{3D}_{t+s_\mathit{out}\cdot n}$. UU is pre-trained on the large motion capture dataset AMASS \cite{amass} and fine-tuned on the target dataset. We adapt this model to estimate the root-relative 3D joint locations and rotations.

\textbf{Rotation Definition.} In this work, we use the same rotations as the SMPL-X body model \cite{smplx}. It consists of 22 joints with a root joint at the pelvis. Each rotation is defined relative to its parent joint. The rotation of the root joint itself is defined relative to the global coordinate system, hence it defines the global rotation of the body. 
We further add 2 joints per hand pose to our set of rotations to capture the position of the hands in more detail. We do not want to estimate the rotations for all fingers, since sports analysts are mainly interested in the body pose. Hence, our main set of rotations consists of 26 joints. However, the set of rotations (called body pose in SMPL-X) can be defined differently depending on the application. For some of our methods, arbitrary rotation definitions are possible, but we also experiment with the SMPL-X body model as an intermediate layer, which only allows SMPL-X compatible rotations.

\subsection{Rotation Representations and Losses}

3D rotations can be represented in various ways. This paper examines three representations: rotation matrices, quaternions, and axis-angle forms. Additionally, we investigate different loss functions for learning rotations. We apply them to both the central output $\tilde{p}^\text{3D}_{t}$ and the output sequence $\mathcal{P}^\text{3D}$, where the mean over all sequence elements is used.

\textbf{Rotation Matrices.} Mathematically, 3D rotations in Euclidean space are represented as rotation matrices $R$ in the special orthogonal group $R \in SO(3) \subset \mathbb{R}^{3 \times 3}$. All these matrices are orthogonal and have a determinant of 1. Using rotation matrices as the network output can not be applied directly, since a neural network can not be constrained to directly output valid rotation matrices. We solve this by projecting the network output to the closest valid rotation matrix regarding the Frobenius norm with a Singular Value Decomposition \cite{levinson2020analysis}.

\textbf{Axis-Angle Form.} The axis-angle representation of a 3D rotation is given by a rotation axis $\omega \in \mathbb{R}^3$ and an angle $\alpha \in \mathbb{R}$, whereby the rotation axis $\omega$ is a unit vector. Axis-angle is more compact than rotation matrices, but it faces the problem of double cover \cite{zhou2019continuity}. This means that it has two representations for the same rotation, which leads to discontinuities in the representation space: It is possible that the shortest distance between two elements in $SO(3)$ corresponds to a much larger distance in the axis-angle representation space which can hinder gradient-based optimization.
Additionally, axis-angle representations can suffer from singularities when the rotation angle approaches zero. 

\textbf{Quaternions.} Quaternions extend the concept of complex numbers to higher dimensions and can be defined through four real values as $q = (w, x, y, z) \in \mathbb{R}^4$.  The rotation axis is defined by the vector $x\mathbf{i} + y\mathbf{j} + z\mathbf{k}$, whereby $\mathbf{i} = (1, 0, 0)$, $\mathbf{j} = (0, 1, 0)$, and $\mathbf{k} = (0, 0, 1)$ are unit vectors. To define the rotation around this axis by an angle $\alpha$, the quaternion $q$ is defined as

\begin{equation}
	q = \cos\left( \alpha / 2\right) + \sin\left( \alpha / 2\right)(x\mathbf{i} + y \mathbf{j} + z\mathbf{k})
\end{equation}

Despite being a robust, efficient, and numerically stable way to handle rotations in 3D space, quaternions also double cover $SO(3)$ \cite{huynh2009metrics}. The quaternions $q$ and $-q$ represent the same rotation in $SO(3)$. As mentioned before, this leads to discontinuities in the representation space and can impede gradient-based optimization.

\begin{figure}[tb]
    \centering
    \includegraphics[width=0.7\linewidth]{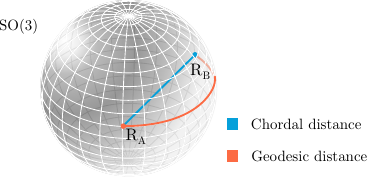}
    \caption{Visualization of the Chordal and Geodesic distances on the $SO(3)$ sphere for two rotation matrices $R_A$ and $R_B$. \cite{losses}}
    \label{fig:loss}
    \vspace{-0.4cm}
\end{figure}
\textbf{Mean Squared Error (MSE).} MSE is a common loss function, which we also use in this work. We apply it to the rotations in the used rotation representation space (rotation matrix, axis-angle or quaternions). For rotation matrices, the MSE loss can be interpreted as the squared Chordal distance \cite{hartley2013rotation}, which is visualized in Figure \ref{fig:loss}. 

\textbf{Geodesic Loss.} The Geodesic loss is calculated only for rotation matrices $R$. In case of another selected representation, we convert it to the corresponding rotation matrix first. Essentially, the Geodesic distance represents the minimal angular difference between rotations. It is defined as:

\begin{equation}
    L_{geo}(R, \tilde{R}) = \frac{1}{K} \sum_{k=1}^K \arccos \left( \frac{\text{trace}(R_k \tilde{R}_k^T) - 1}{2} \right),
\end{equation}
where $K$ is the number of joint rotations, and $\tilde{R}$ and $R$ are the predicted and ground truth rotation matrices, respectively \cite{zhou2019continuity}.
In the case of complete alignment, $L_{geo}$ is zero. Figure \ref{fig:loss} provides a visual interpretation of the Geodesic distance and highlights its difference from the Chordal distance. Visually, the Geodesic distance is the shortest path between two points on the surface of the $SO(3)$ sphere.

\subsection{Fully Supervised Rotation Estimation}\label{sec:supervised}

\begin{figure}[b]
    \centering
    \vspace{-0.2cm}
    \includegraphics[width=0.9\linewidth]{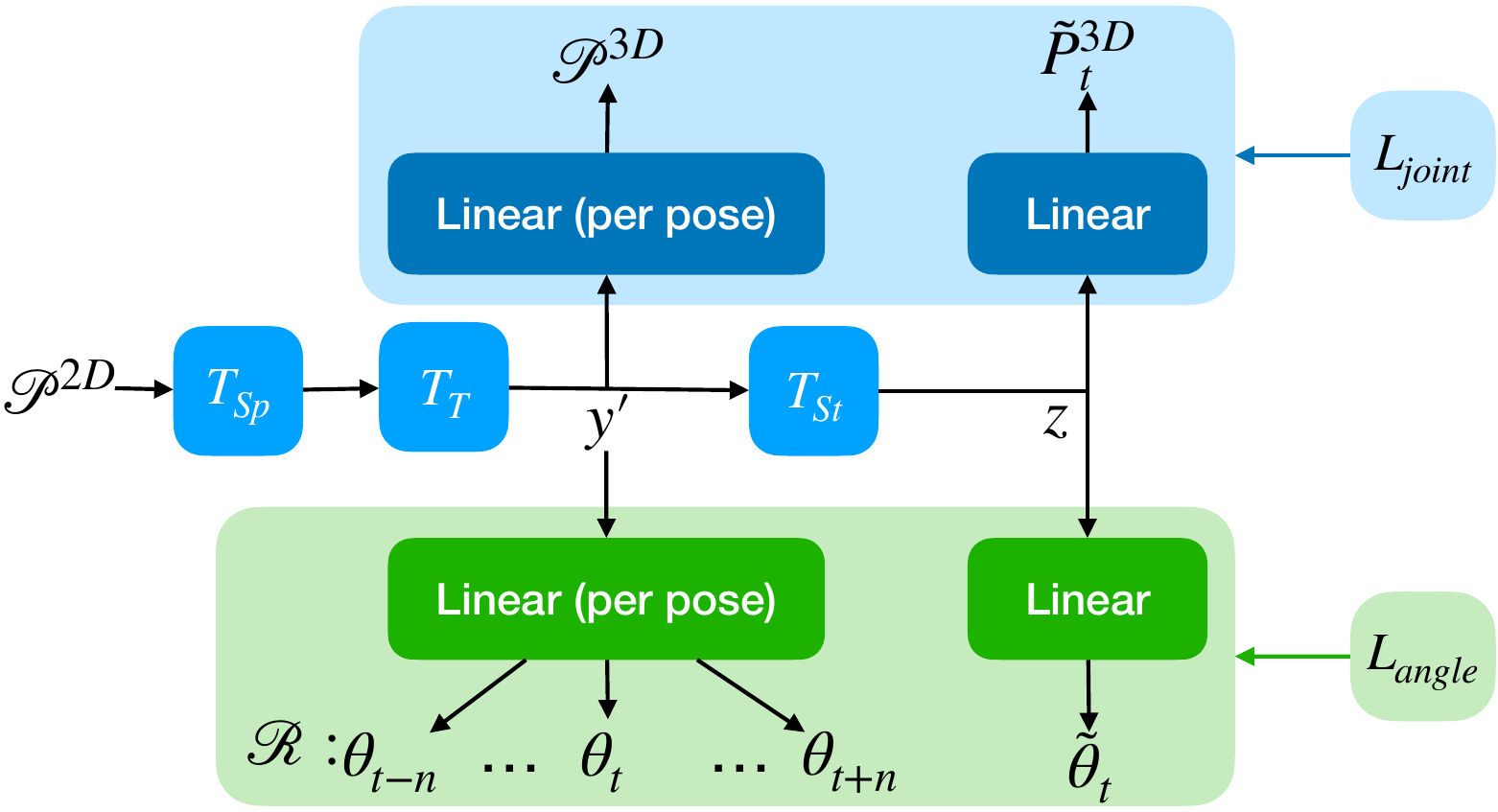}
    \caption{Naive approach to estimate rotations. A second head branch (green) is added to the UU model (blue) to estimate a rotation sequence $\mathcal{R}$ and a refined rotation estimate for the central frame $\tilde{\theta}_t$.}
    \label{fig:naive}
\end{figure}
At first, we introduce model variants that are trained with direct supervision on joint rotations. This is only possible if ground truth rotations are available. We explore possibilities without access to the ground truth in the next section.

\subsubsection{Naive Approach}\label{sec:naive}

In our first approach, we naively extend the UU model to estimate rotations by adding a second head branch for rotation estimation analogous to the joint location estimation. A linear layer is applied to the output of the temporal Transformer $T_T$ to obtain the rotation estimates for the full sequence $\mathcal{R}$. The output of the strided Transformer $T_{St}$ is fed through a linear layer to obtain a refined estimate for the rotations for the central frame $\tilde{\theta}_t$. The loss function $L_\mathit{angle}$ is applied in addition to $L_\mathit{joint}$ to the output sequence $\mathcal{R}$ and the output for the central frame $\tilde{\theta}_t$.
This approach is visualized in Figure \ref{fig:naive}. We combine this approach with all three rotation representations and both loss functions in our experiments.

\subsubsection{Rotation Estimation with a SMPL-X layer} \label{sec:smplx}

In the naive approach, joint rotation and location estimations are completely separate. Since they are actually highly correlated, we propose to unify both estimations by using the SMPL-X body model as an intermediate layer. To use it, we further need body shape parameters $\beta$. Since it is not the focus of this paper, we do not estimate the body shape. During training, we use the ground truth body shape. For evaluation, we further use the A2B methods presented by Ludwig et al. \cite{a2b} to obtain a consistent body shape. Note that our methods combined with such estimated $\beta$ parameters form an HMR method which leverages 2D pose sequences and outperforms the SOTA model Multi-HMR \cite{multihmr} in our experiments (see Section \ref{sec:sota}).

\begin{figure}[b]
    \centering
    \includegraphics[width=\linewidth]{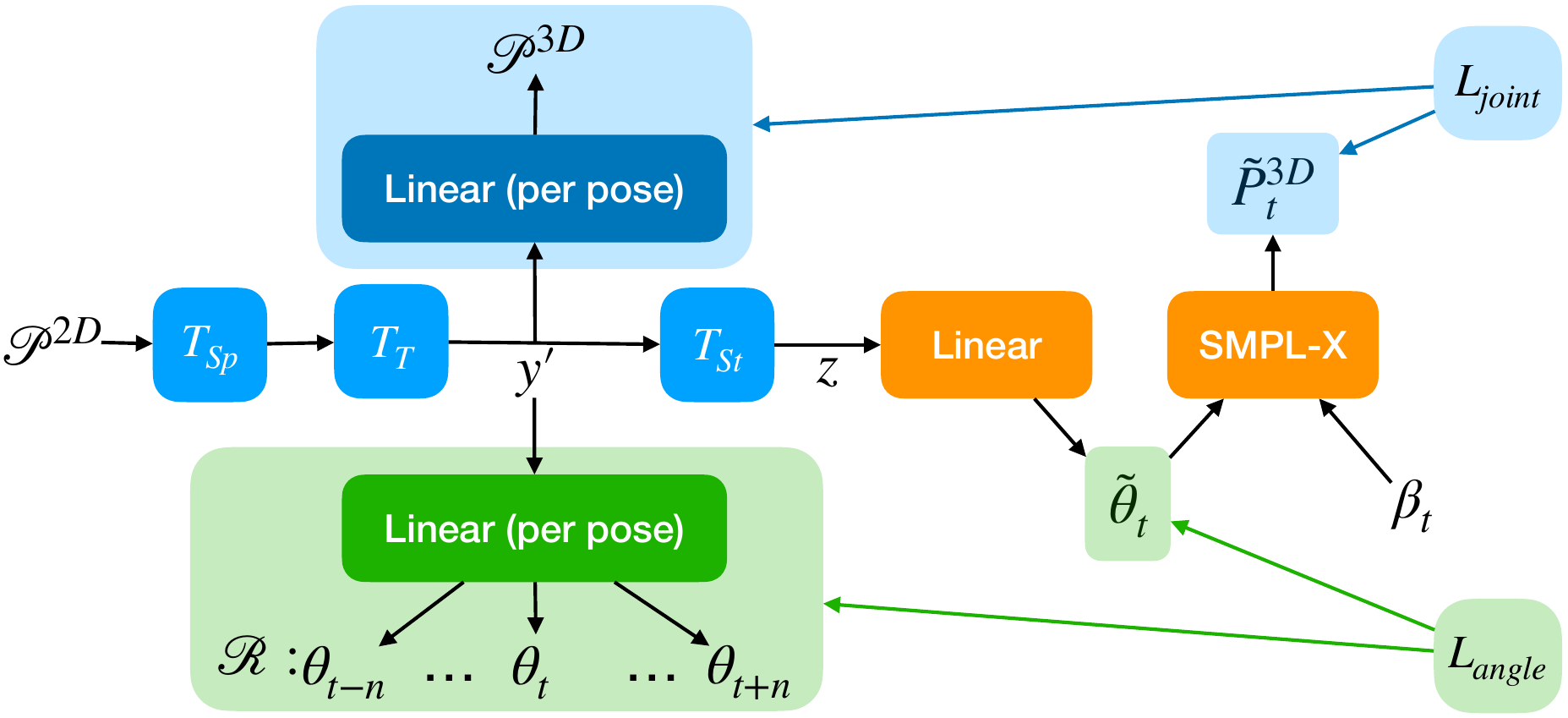}
    \caption{Unified central frame joint rotation and location estimation. The estimated central rotations $\tilde{\theta}_t$ are fed through a SMPL-X layer (orange) to estimate the joint locations. The body shape $\beta_t$ is either the ground truth or obtained with the methods from \cite{a2b}.}
    \label{fig:smplx_sup}
\end{figure}
\begin{figure*}[htb]
    \centering
    \includegraphics[width=0.7\linewidth]{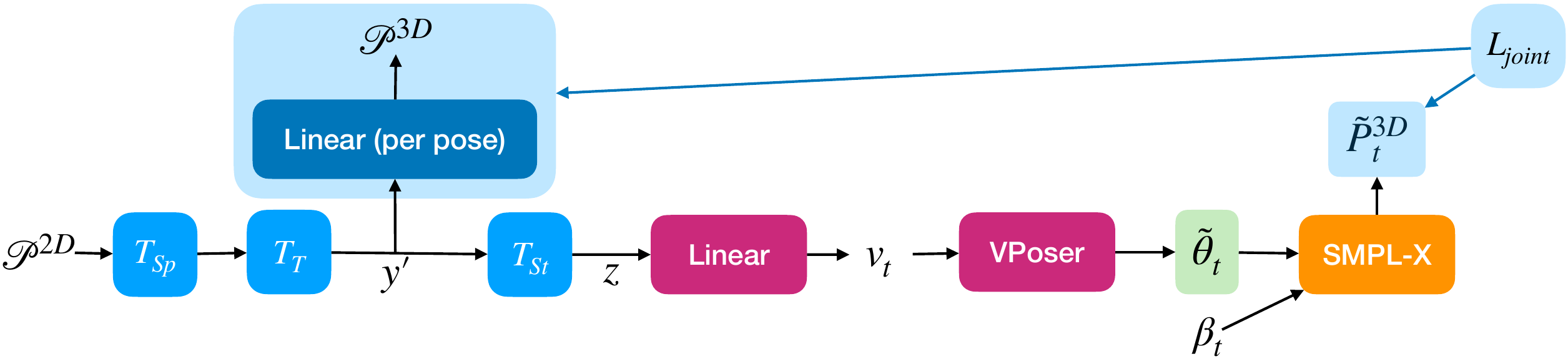}
    \caption{Weakly supervised rotation estimation with the VPoser body prior. The rotation estimation is done in the latent space of VPoser (purple). The estimated rotations are decoded to the joint rotations $\theta_t$ and used to regress the joint locations. Rotations are only estimated for the central frame and indirectly supervised via $L_\mathit{joint}$.}
    \label{fig:vposer}
    \vspace{-0.4cm}
\end{figure*}
The SMPL-X approach consists of a single head branch for the central frame. Since our goal is improved runtime, we refrain from applying a SMPL-X layer to the output sequence since it slows down training and inference. Hence, the model for the output sequence is identical to the naive approach, while the output for the central frame is generated by passing the output $z$ of the strided Transformer $T_{St}$ to a linear layer to estimate $\tilde{\theta}_t$. Next, $\tilde{\theta}_t$ is used as an input to the SMPL-X layer to regress the joint locations. The loss functions $L_\mathit{angle}$ and $L_\mathit{joint}$ are applied to the respective outputs. This approach is visualized in Figure \ref{fig:smplx_sup}.

\subsection{Inverse Kinematics (IK)}\label{sec:ik}

One option to obtain joint rotations $\theta$ and body shape $\beta$ from 3D joint locations only are Inverse Kinematics (IK). Pavlakos et al. \cite{smplx} provide an IK algorithm tailored to the SMPL-X body model. However, since IK involves solving an optimization problem for each frame independently, it is computationally intensive. Ludwig et al. \cite{a2b} utilized IK on outputs from the original UU model to extract joint rotations. We use their model as a baseline and compare our methods to this approach in our experiments.

\textbf{Pseudo Label Approach.} Additionally, IK can be applied to ground truth joint positions to generate pseudo ground truth joint rotations for datasets lacking SMPL-X annotations. This enables the training of fully supervised models using these pseudo labels. 

\subsection{Weakly Supervised Rotation Estimation}\label{sec:weakly}

We further explore alternative models for datasets without access to ground truth rotations. Since the methods still require 3D joint location annotations, we refer to these approaches as weakly supervised.

\subsubsection{Rotation Estimation with a SMPL-X layer}

This approach is very similar to the supervised SMPL-X approach presented in Section \ref{sec:smplx}. The main difference is that $L_\mathit{angle}$ can not be calculated since the necessary ground truth is not available. Hence, the loss function is only $L_\mathit{joint}$ and the central joint rotations are supervised indirectly through the SMPL-X layer. However, there is no supervision of the rotation output sequence $\mathcal{R}$.

\begin{figure}[b]
    \centering
    \vspace{-0.3cm}
    \includegraphics[width=\linewidth]{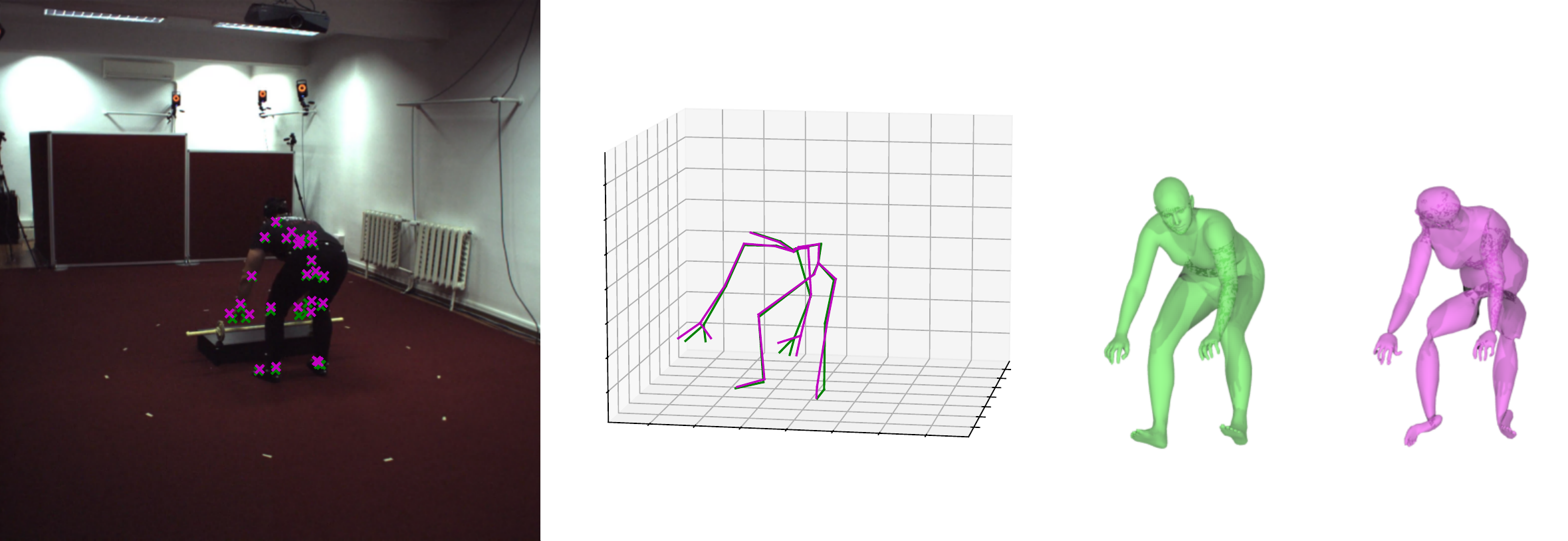}
    \caption{Example prediction for rotation estimation with a SMPL-X layer in weakly supervised manner. The ground truth is displayed in green, the prediction in pink. Joint locations are shown in the first two images, the resulting meshes afterwards. The predicted rotations are completely twisted, resulting in an impossible mesh.}
    \label{fig:smplx_weak}
    \vspace{-0.1cm}
\end{figure}

Experiments show that despite achieving good joint localization accuracy, the joint rotations of this model deviate significantly from the expected output. Since the rotations of the SMPL-X body model are not limited to the same range as the human body, the model learns to predict unrealistic rotations, although the joint locations are fairly accurate. We visualize this problem in Figure \ref{fig:smplx_weak}. Therefore, we will not include this approach in the experiments.

\subsubsection{Rotation Estimation with a Human Body Prior}\label{sec:vposer}

To address this issue, we include a building block that enforces realistic rotations during training. We choose VPoser, which is a human body prior that has learned plausible poses from the large AMASS \cite{amass} dataset. VPoser is an autoencoder that encodes joint rotations $\theta$ into a lower-dimensional latent space. The original SMPL-X $\theta$ parameters for the body pose (without the hands) in an axis-angle format are described by a 63-dimensional vector (3 values for each of the 22 joints apart from the root joint). In contrast, the VPoser latent space has 32 dimensions, and it is normally distributed. This means that the closer a VPoser latent vector is to zero, the more probable is the pose. 

We incorporate VPoser in our model. The output of the strided Transformer $T_{St}$ is fed through a linear layer to estimate the rotations $v_t$ in the latent space of VPoser. VPoser is used to decode $v_t$ to the SMPL-X body pose $\theta_t$. Next, the SMPL-X body model is used as before to regress the joints. The loss function $L_\mathit{joint}$ is applied to the regressed joint locations and the output pose sequence $\mathcal{P}^\text{3D}$. This approach is visualized in Figure \ref{fig:vposer}.

%% file: sec/4_experiments.tex
\section{Experiments}

In this section, we present the experimental results of our proposed model variants. We evaluate \textbf{joint location performance} as well as \textbf{joint rotation performance}. We compare our model variants to the approach involving IK by Ludwig et al. \cite{a2b} and the SOTA HMR model Multi-HMR \cite{multihmr}. Moreover, we evaluate the computational efficiency. For all experiments, we initialize the UU model with the pretrained weights from the AMASS dataset as in \cite{uu}.

\subsection{Evaluation Metrics}

We evaluate the joint rotation performance with the Mean Per Joint Angular Error (MPJAE) \cite{von2018recovering}. Let the relative rotation of joint $k$ be defined by the rotation matrix ${R}_k \in \mathbb{R}^{3\times3}$. For the entire set of $K$ joints, it is represented as ${R} = ({R}_1, \dots, {R}_K)^{T} \in \mathbb{R}^{K\times3\times3}$. The MPJAE measures the geodesic distance between the estimated joint rotations $\tilde{R}$ and the ground truth rotations ${R}$.

For each joint $k$, we define ${R}'_k = {\tilde{R}}_k {R}_k^{T}$. The matrix ${R}'_k$ equals the identity matrix $\mathbf{I} \in \mathbb{R}^{3\times3}$ if the estimate and ground truth perfectly match. Otherwise, ${R}'_k$ is the rotation required to align the predicted orientation with the ground truth. To find the angle of this rotation $\varphi$, we derive it from the trace of the matrix ($\text{trace}({R}'_k) = 1 + 2 \cos \varphi$), using the arc-cosine. This metric reports the error in radians, but we convert it to degrees for easier interpretation in our evaluation tables.
\begin{equation}
	\text{MPJAE}({R}, {\tilde{R}}) = \frac{1}{K} \sum_{k=1}^{K} \arccos \left( \frac{\text{tr}({R}'_k) - 1}{2} \right)
\end{equation}

Moreover, the joint location performance is evaluated using the very common root-relative Mean Per Joint Position Error (MPJPE). 

\subsection{Dataset}

We evaluate our model variants on the \textbf{fit3D dataset} \cite{fit3d}, as it is the only publicly available sports dataset with SMPL-X annotations.
The fit3D dataset comprises videos of human subjects performing various fitness exercises, specifically designed for studying repetitive human motion in a fitness context. The dataset includes recordings of eleven individuals captured in a controlled studio environment from multiple viewpoints, with only one person visible per video. 
The recording setup utilized four synchronized RGB cameras along with a VICON motion capture system consisting of twelve motion cameras. Additionally, each subject was 3D scanned, which is important for gathering the body shape coupled with the body pose.
The recorded exercises target different muscle groups, covering a total of 47 exercises performed by either certified fitness instructors or trainees with varying skill levels.

Fit3D provides official training and test splits. Since the ground truth data is needed for our evaluation to obtain the body shape, and it is only available for the training subset, we do not use the provided test subset in our work. Instead, we divide the original training subset as follows: 6 subjects (s03, s04, s05, s07, s08, s10) are used for training, and 1 subject each for validation (s09) and test (s11).
Since sports analysts focus mainly the body and not the hands/face, we select a subset of joints that we use for our trainings and evaluations. We choose the 22 main body joints and 2 joints on each hand (thumb and pinky). VPoser is only trained on the main body pose, therefore we evaluate both on the 22 main body joints and our 26 joints. 

\subsection{Within-Batch Augmentation}

We use horizontal flipping to augment paired image and 3D pose data in our trainings. We use it in the form of within-batch augmentation (WBA), since it has shown promising results for 2D to 3D uplifting methods, including UU \cite{uu}. WBA means that half of each batch is made up of the original pose sequence and the other half of horizontally flipped pose sequences. 

\subsection{Fully Supervised Training Results}

\textbf{Uplift Upsample.} For comparison, we provide the results of the original UU model. Hence, it is only trained on the 3D joint locations and can not estimate joint rotations. Results are provided in Table \ref{tab:naive}, model {UU}.

\begin{table}[b]
	\centering
    \vspace{-0.2cm}
    \resizebox{0.88\linewidth}{!}{ 
    \begin{tabular}{c|cc|cc}
        \toprule
        Model & $L_\mathit{angle}$ & {WBA} & {MPJPE $\downarrow$} & {MPJAE $\downarrow$}\\
        \midrule
        UU-1 & - & - & 35.51  & - \\
        UU-2 & - & \checkmark & 34.68 & - \\
        \midrule
        N-AA-1 & MSE & - & 62.52 & 10.29   \\
        N-AA-2 & Geodesic & - & \underline{48.35} & 9.39  \\
        N-AA-3 & MSE & \checkmark & 86.18 & 9.38   \\
        N-AA-4 & Geodesic & \checkmark & 49.69  & \underline{9.28} \\
        \midrule
        N-Q-1 & MSE & - & \underline{78.51} & \underline{10.60} \\
        N-Q-2 & Geodesic & - & 563.84 & 85.35  \\
        N-Q-3 & MSE & \checkmark & 95.42 & 11.17   \\
        N-Q-4 & Geodesic & \checkmark & 531.82 & 88.40  \\
        \midrule
        N-RM-1 & MSE & - & 45.66 & 9.33  \\
        N-RM-2 & Geodesic & - & \textbf{39.40} & \textbf{8.82} \\
        N-RM-3 & MSE & \checkmark & 45.05 & 9.21  \\
        N-RM-4 & Geodesic & \checkmark & 41.11 & 8.84  \\
        \bottomrule
    \end{tabular}
    }
    \caption{Results for fully supervised training with UU and the naive method (denoted with \emph{N-}) with all three rotation representations (axis-angle \emph{AA}, quaternions \emph{Q}, and rotation matrices \emph{RM}) and both different loss functions. MPJPE results are given in mm, MPJAE results in degrees. The best results for each model are underlined, the best overall results are marked in bold.}
    \label{tab:naive}
    \vspace{-0.1cm}
\end{table}

\textbf{Naive Approach.} We combine our naive model as explained in Section \ref{sec:naive} with all rotation representations and all loss functions. Results are presented in Table \ref{tab:naive}. We mark all models based on the naive approach with a leading \emph{N-} and name them according to their rotation representation \emph{AA} for axis-angle, \emph{Q} for quaternions, and \emph{RM} for rotation matrices. We observe very different results for the different representations. Quaternions perform badly, while axis-angle and rotation matrices show promising results, especially with geodesic loss, while quaternions perform better with MSE loss. 
Interestingly, WBA fails to improve the naive model's results, unlike observed with the UU model.
The best results are achieved with rotation matrices and geodesic loss.

\begin{table}[tb]
	\centering
    \resizebox{0.85\linewidth}{!}{ 
    \begin{tabular}{c|cc|cc}
        \toprule
        Model & $L_\mathit{angle}$ & {WBA} & {MPJPE} & MPJAE\\
        \midrule
        S-AA-1 & MSE &  - & 62.48 & \textbf{9.14} \\
        S-AA-2 & Geodesic  & - & 54.03  & 9.72 \\
        S-AA-3 & MSE  & \checkmark & \textbf{36.69}  & 9.21 \\
        S-AA-4 & Geodesic & \checkmark & 41.31  & 9.23 \\
        \midrule
        S-RM-1 & MSE & - & 42.90 & 9.42 \\
        S-RM-2 & Geodesic  & - & 41.90  & 9.44 \\
        S-RM-3 & MSE  & \checkmark & 42.69 & 9.22\\
        S-RM-4 & Geodesic  & \checkmark & \underline{40.21}  & \underline{9.19} \\
        \bottomrule
    \end{tabular}
    }
    \caption{Results for fully supervised training with the SMPL-X layer method (denoted with \emph{S-}) for axis-angle \emph{(AA)} and rotation matrix \emph{(RM)} rotation representations and both loss functions. MPJPE results are given in mm, MPJAE results in degrees. The best results for each model are underlined, the best overall results for model variants with rotation estimation are marked in bold.}
    \label{tab:smplx_sup}
    \vspace{-0.4cm}
\end{table}

\textbf{SMPL-X Layer Approach.} Next, we evaluate the approach including an SMPL-X layer (see Section \ref{sec:smplx}). We combine it with both loss functions and rotation matrices and axis-angle rotation representations. We leave out quaternions since they perform so badly in the naive approach and since they are not used in the SMPL-X model itself as representations. Results are presented in Table \ref{tab:smplx_sup} and marked with a leading \emph{S-}. The results differ from the naive approach. Axis-angle now outperform rotation matrices. The best MPJAE result is achieved with MSE and without WBA (S-AA-1), but the MPJPE of this experiment is relatively bad, over 28\,mm higher than the original UU model. However, with WBA (S-AA-3), the MPJAE rises a little, but the MPJPE is reduced to 36.69\,mm, which is only 2\,mm higher than the original UU model and nearly 3\,mm lower than the best naive model. Rotation matrices achieve similar MPJAE scores, but the best MPJPE score is over 3\,mm higher than the best axis-angle model. Qualitative results of the overall best supervised model S-AA-3 are shown in Figure \ref{fig:vis}.

\subsection{Weakly Supervised Training Results}

\textbf{Pseudo Label Approach.} We run IK on the ground truth 3D joint locations to obtain pseudo labels for the joint rotations. This way, we do not use the ground truth joint rotations but can use the fully supervised training routines (see Section \ref{sec:ik}, pseudo label approach). We use the IK routine provided by Pavlakos et al. \cite{smplx}, which estimates only the main body pose and not the hand pose. Therefore, we also evaluate only on the 22 joints of the main body pose. We show the result of this experiment with the same two settings as the best supervised models (S-AA-1 and S-AA-3) in Table \ref{tab:weakly}, model \emph{PS-AA-1/3}. 

\textbf{VPoser Approach.} We use VPoser as a human body prior to prevent the weakly supervised model from estimating unrealistic poses (see Section \ref{sec:vposer}). Results of this experiment are shown in Table \ref{tab:weakly}, with models marked by a leading \emph{V-}. 
VPoser is trained only on the main body pose, so we evaluate on the 22 main body joints. Results for the 26-joint set for the best models are presented in Table \ref{tab:comp}.

\begin{table}[tb]
	\centering
    \resizebox{0.75\linewidth}{!}{ 
    \begin{tabular}{cc|cc}
        \toprule
        Model & WBA & MPJPE & MPJAE-22 \\
        \midrule
        PS-AA-1  & - & 38.05 & 16.20  \\
        PS-AA-3  & \checkmark & \textbf{37.30} & \underline{16.18}\\
        \midrule
        V-1 & - & 40.40 & 16.11 \\
        V-2 & \checkmark & \underline{38.76} & \textbf{15.90}\\
        \bottomrule
    \end{tabular}
    }
    \caption{Results for models with joint location annotations only (MPJPE in mm, MPJAE in degrees). The SMPL-X layer method with pseudo labels is marked with leading \emph{PS-AA-}, the VPoser approaches with \emph{V-}. Best results for each model are underlined, the best overall results are bold.}
    \label{tab:weakly}
    \vspace{-0.3cm}
\end{table}

The experiments show that the MPJAE is generally much higher than with direct supervision, which is something to be expected. The best weakly supervised model regarding the MPJAE is the VPoser model V-2. Compared to the best supervised model, the MPJPE scores for the weakly supervised models are only slightly higher and even better than the best naive model, most likely because joint locations are supervised with ground truth labels. The MPJPE results are best for the pseudo label approach PS-AA-3. Since V-2 achieves a little better MPJAE and PS-AA-3 a little better MPJPE, there is not a single best model in the weakly supervised case. However, the best results are achieved with WBA in both cases.

\begin{table*}[htb]
	\centering
    \resizebox{0.75\linewidth}{!}{ 
    \begin{tabular}{cc|cccc|c}
        \toprule
        Model & GT rot. & MPJPE-26 & MPJAE-26 & MPJAE-22 & MPJPE-37 & Runtime [ms]\\
        \midrule
        UU-2 & - & 34.68 & - & - & 34.3 \cite{a2b} & 7.11 $\pm$ 2.23 \\
        \midrule
        \midrule
        N-RM-2 & \checkmark  & {39.40} & \textbf{8.82} & \textbf{8.79} & 42.38 &\textbf{9.86 $\pm$ 2.78}\\
        S-AA-3 & \checkmark  & {\textbf{36.69}}  & 9.21  & 9.24 & 41.14 &11.09 $\pm$ 3.14\\
        \midrule
        PS-AA-3  & - & {37.30} & (18.09) &{16.18} & 42.33 &\textbf{11.46 $\pm$ 3.50}\\
        V-2 &      - & {38.76} & (17.86) &\textbf{15.90} & 44.43 &16.08 $\pm$ 4.24\\
        \midrule
        Multi-HMR \cite{multihmr} & (\checkmark) & 62.22 & 18.49 & 17.59 & 68.96 & 156.10 $\pm$ 10.09\\
        \multirow{2}{*}{UU-IK \cite{a2b}} & \multirow{2}{*}{-} & \multirow{2}{*}{\textbf{34.62}} & \multirow{2}{*}{(18.30)} & \multirow{2}{*}{16.43} & \multirow{2}{*}{{36.91}} & 1952.19 $\pm$ 1091.52 / \\
        &&&&&& 4733.31 $\pm$ 1826.99\\
        \bottomrule
    \end{tabular}
    }
    \caption{Results of our best models and other SOTA models based on ground truth body shape (MPJPE in mm, MPJAE in degrees). The number of joints involved in the metric calculation is denoted after the respective metric name. The original UU model is displayed only for comparison and not included in highlighting best models since it is not capable of estimating rotations. We highlight best results (MPJPE, MPJAE and runtime) for models trained with and without ground truth rotations (GT rot.) available during training. Some models do not provide rotations for fingers. In these cases, we set them to 0 and put MPJAE-26 results in brackets. Since Multi-HMR is trained with rotation supervision, but not on the fit3D dataset, we put the checkmark in brackets. The runtime is measured in ms. For UU-IK, we provide the runtime for frames with (first value) and without (second value) pre-initialization.}
    \label{tab:comp}
    \vspace{-0.3cm}
\end{table*}

\subsection{Comparison with other SOTA Methods and Runtime Evaluation}\label{sec:sota}

After evaluating our own model variants and identifying the best versions, we compare them with other SOTA methods. On the one hand, we choose Multi-HMR \cite{multihmr}, which is a SOTA model for HMR and showed the best performance on fit3D \cite{a2b}. It operates image-wise and estimates SMPL-X human meshes. In our evaluations, we keep the estimated rotations and combine them with the ground truth body shape to achieve a fair evaluation. On the other hand, we choose the pipeline involving UU and IK (called UU-IK) proposed by Ludwig et al. \cite{a2b}. At first, we also use the ground truth body shape for this model. We evaluate with estimated body shapes in Section \ref{sec:a2b}. Results are displayed in Table \ref{tab:comp}. We include the MPJAE on our full set of 26 joint rotations, also for the models which do not estimate the hand rotations (we leave them in the template pose in these cases). For comparison, we include the MPJPE on the same set of 37 keypoints as selected by Ludwig et al. \cite{a2b} (our set contains 26 keypoints). The reader should keep in mind that these evaluations are biased towards their model, since scores are improving for keypoints included in the training target and nearly a third of them are not included in our losses. Moreover, we provide a runtime evaluation. We measure the time needed for a forward pass on a single image using a NVIDIA GeForce RTX 2080 Ti GPU. We provide the mean and standard deviation based on at least 5k forward passes. For UU-IK we measure the runtime separately for frames initialized with the pose from the previous frame and without pre-initialization (see \cite{a2b}). 

Our evaluations show that \textbf{our best fully supervised models outperform all other models regarding MPJAE scores}. Multi-HMR performs worst regarding MPJPE and MPJAE scores. UU-IK achieves the best MPJPE scores, but the MPJAE scores are worse compared to our models. Its significant problem is further the runtime. While having an average runtime of over 1900\,ms even for frames with pre-initialization, our best models only need 9.86\,ms and 11.48\,ms (best fully and weakly supervised model, respectively). This is over \textbf{150 times faster}, which proves that we achieve our goal of providing a faster and comparably accurate 2D to 3D pose uplifting model including joint rotations. 

\subsection{Evaluation with Consistent Body Shape}\label{sec:a2b}

\begin{table}[tb]
	\centering
    \resizebox{0.8\linewidth}{!}{ 
    \begin{tabular}{cc|cc}
        \toprule
        Model & GT rot. & MPJPE-26 & MPJPE-37 \\
        \midrule
        \midrule
        N-RM-2 & \checkmark  & 41.01 & 43.95\\
        S-AA-3 & \checkmark  & 38.35 & 42.72\\
        \midrule
        PS-AA-3  & - & 39.02 & 43.98\\
        V-2 &      - &  40.39 & 46.01\\
        \midrule
        UU-IK \cite{a2b} & - & 35.71 & 38.41\\
        \bottomrule
    \end{tabular}
    }
    \caption{MPJPE results (for 26 and 37 keypoints) of our best models and other SOTA models based on A2B body shape \cite{a2b} in mm. We only display the result of the best of the four A2B model variants. For MPJAE, see Table \ref{tab:comp}.}
    \label{tab:a2b}
    \vspace{-0.4cm}
\end{table}
Until now, we evaluated with the ground truth body shape. Lastly, we evaluate the best models with an estimated body shape. We use the A2B models suggested by Ludwig et al. \cite{a2b}, which estimate the body shape based on anthropometric measurements. This makes sense in sports, since professional athletes are measured for their analyses. Moreover, we use a single set of estimated body shape parameters to ensure a consistent body shape across all frames of a video, as suggested by \cite{a2b}. Since we used the ground truth body shape so far, the evaluations might not seem completely realistic. However, the fit3D ground truth itself is not consistent as shown by \cite{a2b}, which makes a completely fair evaluation with consistent body shapes impossible. Results are shown in Table \ref{tab:a2b}. Only MPJPE values are included since MPJAE is not affected by the body shape.

%% file: sec/5_conclusion.tex
\section{Conclusion}

In this paper, we have proposed several novel model variants for \textbf{efficiently estimating a full 3D human pose, including joint rotations, based on a 2D pose sequence}. We explored models trained with and without ground truth rotations. We have shown that our models outperform the state-of-the-art in terms of rotation estimation accuracy. Furthermore, we have demonstrated that our models are computationally more efficient than the method of Ludwig et al. \cite{a2b} by eliminating the need for an additional inverse kinematics step. Our models are particularly well-suited for sports analytics, as they provide accurate joint rotations, which are crucial for understanding an athlete's biomechanics.